\pdfoutput=1

\documentclass[11pt]{article}

\usepackage[final]{acl}
\usepackage{hyperref}

\usepackage{times}
\usepackage{latexsym}

\usepackage[T1]{fontenc}

\usepackage[utf8]{inputenc}

\usepackage{microtype}

\usepackage{inconsolata}

\usepackage{graphicx}
\usepackage{ulem}
\usepackage{tcolorbox} 
\usepackage{multirow}
\usepackage{enumitem}
\usepackage{makecell}
\usepackage{algorithmic}
\usepackage{algorithm}
\usepackage{booktabs}
\usepackage{graphicx}
\usepackage{amsmath}
\usepackage{xcolor}
\usepackage{colortbl}
\usepackage{listings}
\title{Beyond Prompt Content: Enhancing LLM Performance via \\ Content-Format Integrated Prompt Optimization}

\author{ 
    Yuanye Liu$^{*\dagger}$ \hspace{3pt}
    Jiahang Xu$^{*\diamond}$ \hspace{3pt}
    Li Lyna Zhang \hspace{3pt}
    Qi Chen \hspace{3pt}
    Xuan Feng \hspace{3pt}\\
    \textbf{Yang Chen} \hspace{3pt}
    \textbf{Zhongxin Guo} \hspace{3pt}
    \textbf{Yuqing Yang} \hspace{3pt}
    \textbf{Peng Cheng} \hspace{3pt}
    \\  \fontsize{10}{10} \selectfont{$^\dagger$Fudan University \hspace{5pt} Microsoft Research Asia}  
}

\begin{document}

\arrayrulecolor{black}
\newcommand{\sysname}{{CFPO}}
\newcommand{\fullsysname}{{Content-Format Integrated Prompt Optimization (CFPO)}}

\newcommand{\lz}[1]{{\textcolor{red}{\it Lyna: #1}}}
\newcommand{\yuanye}[1]{{\textcolor{cyan}{\it Yuanye: #1}}}
\newcommand{\qi}[1]{{\textcolor{purple}{\it Qi: #1}}}

\newcommand{\key}[1]{{\textcolor{blue}{\textbf{[#1]}}}}

\maketitle
\def\thefootnote{$*$}\footnotetext{Equal contribution.}
\def\thefootnote{$\dagger$}\footnotetext{Yuanye Liu did the work during an internship at MSRA.}
\def\thefootnote{$\diamond$}\footnotetext{Corresponding author: jiahangxu@microsoft.com}
\begin{abstract}

Large Language Models (LLMs) have shown significant capability across various tasks, with their real-world effectiveness often driven by prompt design. 
While recent research has focused on optimizing prompt content, the role of prompt formatting—a critical but often overlooked dimension—has received limited systematic investigation.
In this paper, we introduce \fullsysname{}, an innovative methodology that jointly optimizes both prompt content and formatting through an iterative refinement process.
\sysname{} leverages natural language mutations to explore content variations and employs a dynamic format exploration strategy that systematically evaluates diverse format options.
Our extensive evaluations across multiple tasks and open-source LLMs demonstrate that \sysname{} demonstrates measurable performance improvements compared to content-only optimization methods. This highlights the importance of integrated content-format optimization and offers a practical, model-agnostic approach to enhancing LLM performance.
Code is available at 
\href{https://github.com/HenryLau7/CFPO}{https://github.com/HenryLau7/CFPO}.
\end{abstract}

\section{Introduction}

\begin{figure}[h]
    \vspace{-2ex}
    \includegraphics[width=\columnwidth]{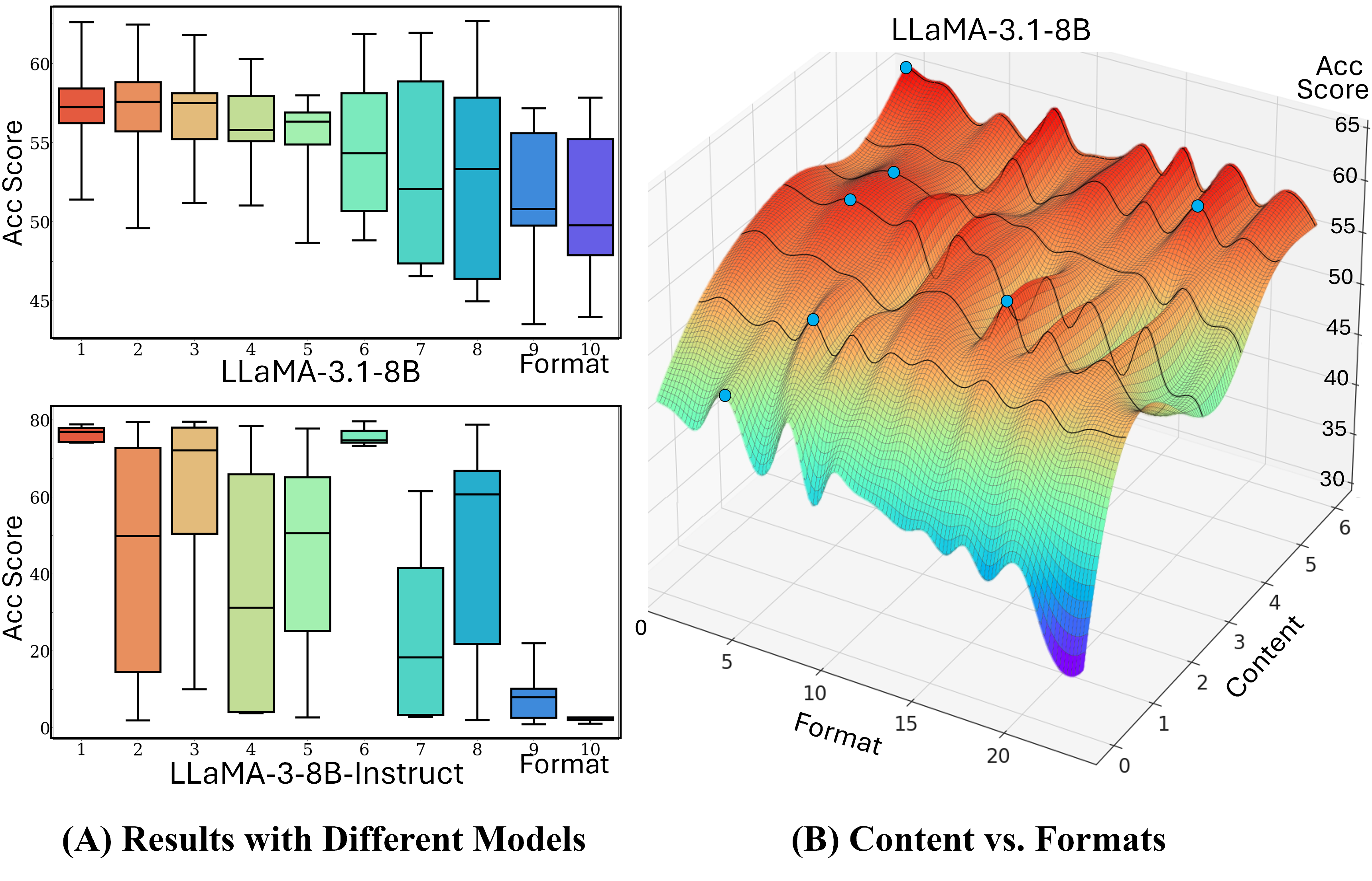}
    \vspace{-3.5ex}
    \caption{
    The crucial role of prompt formatting and its interaction with content. \textbf{(A)}: Model-specific format sensitivity: LLM performance varies significantly across different prompt formats on GSM8K task. \textbf{(B)}: Content-format interdependence: The optimal format for a prompt depends on its content, highlighting the need for joint optimization.}
    \vspace{-3ex}
\label{fig:teaser}
\end{figure}

Large Language Models (LLMs) have demonstrated impressive achievements across various domains~\cite{openai2024gpt4technicalreport}. The effectiveness of LLMs in real-world applications is fundamentally dependent on the design of effective prompts, which serve as an essential interface between human users or developers and the LLM system. Studies have shown that expert-designed prompts could significantly enhance LLM performance~\citep{brown2020incontextlearning, wei2023chainofthought,A_2024_prompt_survey}.

Despite its significance, manual prompt design is fraught with challenges. LLMs are notoriously sensitive to subtle variations in prompt phrasing and structure, with performance often differing markedly across models and tasks based on these nuances~\citep{zhuo-etal-2024-prosa, chatterjee2024posixpromptsensitivityindex, 
Jiang2022promptMaker, salinas2024butterflyeffectalteringprompts, zhuo2024prosaassessingunderstandingprompt, C_2024ICLR_formatspread}. To alleviate these difficulties, automated prompt optimization techniques, often 
leveraging enhanced LLMs to optimize prompts, have proven to be effective to adapt and refine prompts~\cite{C_2023EMNLP_APO, C_2024EMNLP_sammo, C_2024ICLR_LLMasOPT}.

However, existing research primarily focuses on optimizing \textit{prompt content}, while overlooking a critical and unexplored dimension: the \textit{\textbf{prompt formatting}}.
Prompt formatting refers to the arrangement and presentation of prompt content while preserving its semantic meaning. As LLMs are applied to increasingly complex tasks, structuring prompts into distinct components (e.g., instructions, examples, queries) becomes paramount for effectively conveying the desired task and context.  Thus, the manner in which a prompt is formatted can significantly impact performance.

Our preliminary investigations (Figure~\ref{fig:teaser}) highlight the significant impact of prompt format on LLM performance.
We observed that different LLMs exhibit distinct format preferences, with formats performing well on one model sometimes failing on another. This underscores the existence of sophisticated, model-specific format biases~\cite{C_2024ICLR_formatspread}. Furthermore, even within a single LLM, the optimal format varies depending on the specific prompt content. This complex interplay between content and format suggests that a one-size-fits-all approach to prompt formatting is unlikely to succeed, highlighting the need for joint optimization strategies that consider content and format as interdependent variables.

To address these limitations, we introduce \textbf{\fullsysname{}}, an innovative methodology that concurrently optimizes both prompt content and format through an iterative refinement process. \sysname{} employs distinct optimization strategies tailored to the unique search spaces of content and format. Content optimization is guided by performance feedback and Monte Carlo sampling, leveraging natural language mutations to enhance prompt effectiveness. For format optimization, \sysname{} explores a discrete set of format options through a dynamic exploration strategy designed to identify optimal formats without requiring prior knowledge.

Specifically, \sysname{}'s format optimizer draws upon principles of structured thinking and defines a hierarchical template that clearly demarcates content elements from their formatting attributes. This allows for targeted optimization of both intra-component styling (e.g., how an example is presented)~\citep{voronov2024mindformatconsistentevaluation,salinas2024butterflyeffectalteringprompts} and inter-component arrangement (e.g., the order and connectors between components)~\citep{he2024doespromptformattingimpact}, adapting to the specific needs of different prompt components and their interactions.

Our primary contributions are threefold: 

\begin{itemize}[itemsep=-3pt, topsep=0pt, leftmargin=*, itemindent=-\labelsep]
\item We propose \sysname{}, an innovative approach to simultaneously optimizes prompt content and format using an iterative process. 

\item We introduce an efficient strategy for dynamic format optimization that iteratively generates and evaluates format candidates using a scoring system to select superior options.

\item We conduct extensive evaluations across diverse tasks and open-source LLMs, showing that \sysname{} consistently enhances model performance in a measurable and effective way.
\end{itemize}
\section{Related Work}

\begin{figure*}[t]
  \vspace{-2ex}
  \includegraphics[width=\linewidth]{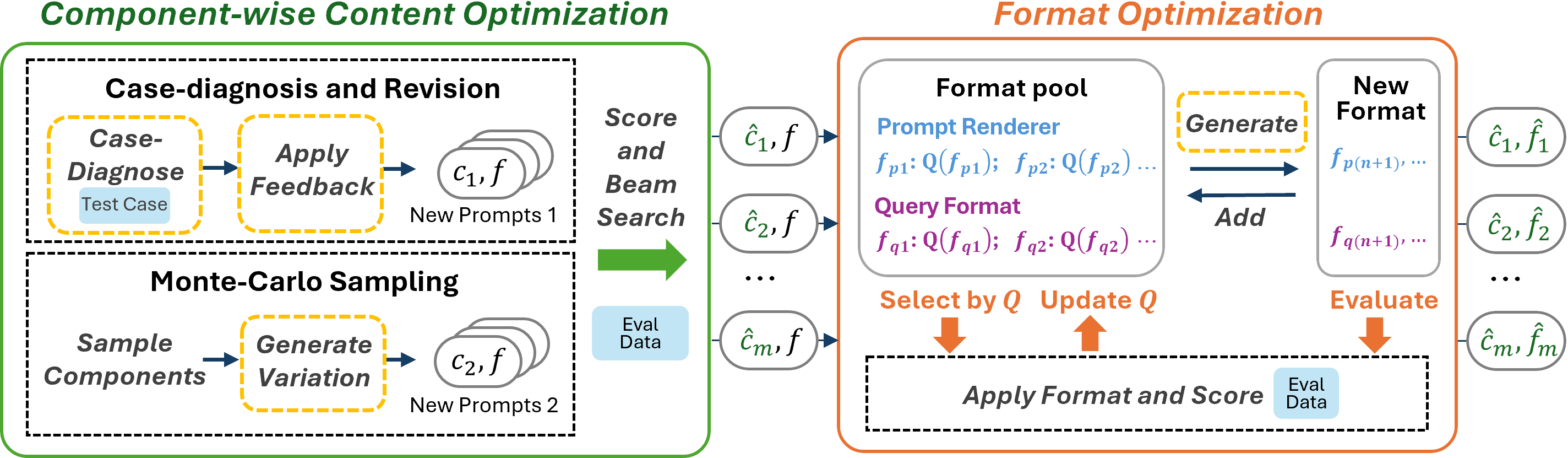}
  \vspace{-3ex}
  \caption {Illustration of the \sysname{} pipeline within a single iteration round. In the initial Component-wise Content Optimization stage, case-diagnosis and Monte-Carlo sampling are employed for content mutation. Subsequently, the Format Optimization stage identifies the most suitable format for each content candidate. The yellow dashed line indicates where the LLM optimizer is employed to guide the optimization process.}
  \label{fig:pipeline}
  \vspace{-2ex}
\end{figure*}


\subsection{Optimization via LLM}
The remarkable capacity of LLMs has been demonstrated in various tasks as optimizers, leveraging their ability to enhance performance, such as code generation~\citep{A_2023_optimize_code, C_2024COLM_stop, A_2024_MAGIC}, tool-making~\citep{A_2024_tool_making}, and agent system design~\citep{hu2024automateddesignagenticsystems}. 
However, recent studies indicate that LLMs face significant challenges in achieving completely automatic optimization. These models often rely on human intervention for designing workflows and struggle with tasks requiring complex decomposition and iterative refinement~\citep{A_2024_AFlow,J_2024TMLR_moreagent}.


\subsection{Prompt Optimization Techniques}
\noindent \textbf{Automatic Prompt Optimization} \label{sec:related_work}
Automatic prompt optimization plays a crucial role in enhancing the performance of LLMs by refining prompts without requiring human intervention. 
Various approaches have been explored to search for the optimal prompt, including reinforcement learning~\citep{C_2023ICLR_TEMPERA}, Monte Carlo Search~\citep{zhou2023ape}, Monte Carlo Tree Search (MCTS)~\citep{C_2024ICLR_promptagent}, and agent-driven frameworks~\citep{A_2024_LangGPT,C_2024ICLR_DSPy,O_2023_Autogpt}. 
Notably, feedback-based methods have emerged as a significant approach~\citep{C_2023EMNLP_APO, A_2024_GReaTer}. These methods iteratively evaluate prompts on a batch of evaluation data, using error cases to guide subsequent mutation and improvement \citep{C_2023EMNLP_APO, A_2024_GReaTer}. In each iteration, the prompt optimizer evaluates the prompt on the evaluation set and feeds the incorrectly predicted test cases back to the optimizer to guide improvement in the next mutation step.
However, existing automatic prompt optimization techniques often lack the capacity for fine-grained modifications. While \citep{C_2024ICLR_DSPy,C_2024EMNLP_sammo} introduce phrase-level mutations, a systematic approach to optimizing \textit{prompt format} is lacking. This leaves a gap to comprehensively adapt prompt structure for optimal performance.

\noindent \textbf{Prefix Tunning} offers an effective strategy for adapting large language models (LLMs) to specific tasks by learning continuous, task-specific vectors that are prepended to the input sequence~\cite{li2021prefix,wang2023multitaskprompttuningenables,guo2024qtuningqueuebasedprompttuning,gu-etal-2022-ppt,wang2023multitaskprompttuningenables}. While demonstrating impressive performance gains, a key limitation of Prefix Tuning lies in its reliance on access to the LLM's internal parameters for training.  This requirement presents a significant obstacle when working with black-box LLMs accessed through APIs or in resource-constrained environments where full fine-tuning is infeasible.

\subsection{Prompt Structure and Sensitivity}
\noindent \textbf{Structured Prompting}, which organizes prompts into components like instructions and examples, is a widely recommended practice \citep{openai_guide,google2024Promptingguide101}. Frameworks like LangGPT \citep{A_2024_LangGPT} further advance this paradigm by introducing reusable prompt designs inspired by software engineering principles, showcasing the potential of structured approaches \citep{promptbreeder}.  While these efforts have primarily focused on optimizing the content and organization of prompt components, less attention has been paid to the impact of \textit{prompt formatting} within this structured context.

\noindent \textbf{Sensitivity of Prompt Variations} 
LLMs exhibit significant sensitivity to prompt variations. Studies have shown that even semantically similar but unseen instructions can lead to performance degradation~\citep{sun2023evaluatingzeroshotrobustnessinstructiontuned, mizrahi2024stateartmultipromptllm}. To address this brittleness, researchers have proposed methods and metrics to systematically evaluate and understand prompt sensitivity~\citep{zhuo2024prosaassessingunderstandingprompt, chatterjee2024posixpromptsensitivityindex}. These findings underscore the necessity of prompt optimization for robust LLM performance.

\noindent \textbf{Format Sensitivity of Prompt}
Studies have highlighted the impact of formatting on prompt performance~\citep{salinas2024butterflyeffectalteringprompts}.
\citet{C_2024ICLR_formatspread} revealed that modifications to separators and spacing within a query could substantially impact performance.
\citet{he2024doespromptformattingimpact}
reveals that the format of prompts significantly impacts GPT-based models’ performance, with no single format excelling universally.
\citet{voronov-etal-2024-mind} focuses on the format of few-shot examples and suggests that it is beneficial to maintain a consistent format across examples.
Despite the growing recognition of formatting's influence, current prompt engineering practices rely heavily on empirical observations and lack systematic design principles.

\section{\sysname{}: Content-Format Integrated Prompt Optimization}
\label{sec:method}

The effectiveness of LLMs is profoundly influenced by both the content and format of prompts. To address this, we propose \fullsysname{}, a novel framework designed to jointly optimize prompt content and structure for enhanced LLM performance. \sysname{}, illustrated in Figure~\ref{fig:pipeline}, employs a dual-optimizer approach in each iteration. 
The subsequent sections provide a detailed explanation of our methodology, beginning with the structured prompt template we utilize (Section~\ref{sec:template}), followed by a description of the Component-wise Content Optimization (Section~\ref{sec:content_optimizer}), the Format Optimization (Section~\ref{sec:format_optimizer}), and finally, the integration of these optimizers into the complete \sysname{} framework (Section~\ref{sec:dual_optimizer}).
Implementation details and meta-prompts are provided in Appendix \ref{apx:meta_prompts}.

\begin{figure*}[t]
  \vspace{-2ex}
  \includegraphics[width=\linewidth]{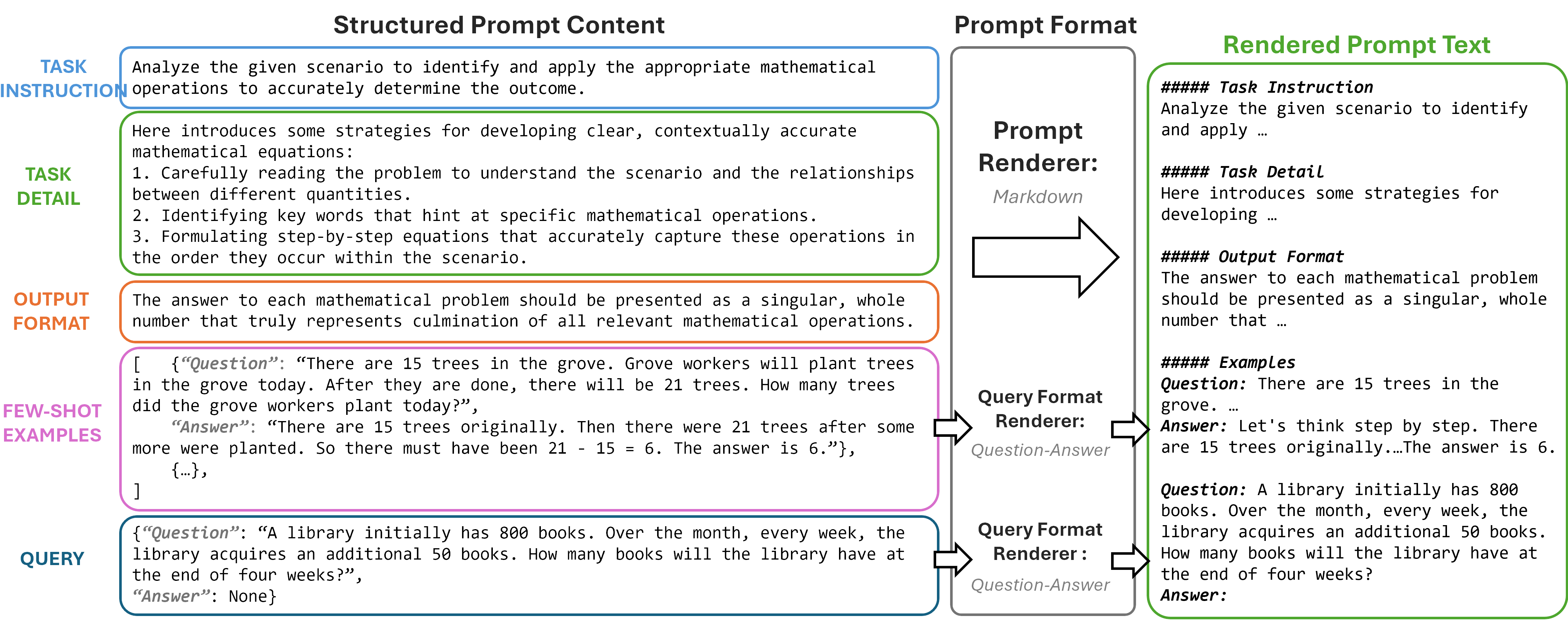}
  \vspace{-4.5ex}
  \caption {An illustrative example of our Structured Prompt Template. This template systematically organizes the prompt into distinct components, each serving a specific functional role. When formulating a prompt, the template first employs a Query format to present examples and queries, and then integrates all content components via the Prompt Renderer to construct the comprehensive prompt string.}
  \vspace{-2ex}
  \label{fig:template}
\end{figure*}

\subsection{Structured Prompt Template}
\label{sec:template}

To facilitate fine-grained and targeted optimization, our framework adopts a structured prompt template inspired by prompt engineering guidelines from ~\citet{openai_guide} and~\citet{google2024Promptingguide101}. Our template decomposes prompts into distinct functional components, allowing for both detailed analysis and selective modification. The template distinguishes between content-based and format-based components, providing flexibility and adaptability to diverse user needs. A set of common and generally applicable elements is illustrated in Figure~\ref{fig:template}.

The \textbf{Content-based Components} define the information provided to the LLM, including:

\begin{itemize}[itemsep=-4pt, topsep=0pt, leftmargin=*, itemindent=-\labelsep]
\item \textbf{Task Instruction} defines the primary goal, guiding the model's overall behavior.

\item \textbf{Task Detail} offers supplementary task-specific information, including resolution steps.

\item \textbf{Output Format} specifies the desired output structure (e.g., JSON, bullet points, etc.).

\item \textbf{Few-shot Examples} provide specific instances for contextual learning patterns. 

\item \textbf{Query} shows the question or request to be answered by the LLM.
\end{itemize}


The \textbf{Format-based Components} dictate how the prompt is assembled and presented, including:
\begin{itemize}[itemsep=-4pt, topsep=0pt, leftmargin=*, itemindent=-\labelsep]

\item \textbf{Prompt Renderer} defines how to aggregate all components into a structured prompt.
\item \textbf{Query Format}: defines how to structure the rendering of examples and queries.

\end{itemize}




Note that this template is highly adaptable. Users can readily adjust the template by adding or deleting components, or incorporating additional formatting elements (e.g., tables, structured documents) within existing components to suit their specific requirements. By decoupling format from content, this structured template empowers users to perform targeted and precise optimizations, leading to improved prompt effectiveness.

\subsection{Component-wise Content Optimization}
\label{sec:content_optimizer}

Building upon existing prompt optimization strategies, which often rely on either feedback-driven refinement or Monte-Carlo sampling (Section~\ref{sec:related_work}), \sysname{} introduces a more targeted and efficient approach to content optimization. While feedback-driven methods diagnose weaknesses through failure analysis, and Monte-Carlo sampling diversifies perspectives with semantic variations, \sysname{} innovates along two key dimensions.

First, \sysname{} expands the diagnostic phase beyond traditional failure analysis by incorporating \textit{correct cases}. This allows the LLM to identify and reinforce successful aspects of the prompt, complementing error correction. Second, \sysname{} adopts a \textit{component-wise optimization strategy}. Instead of treating the prompt as a monolithic block, \sysname{} targets specific content-based components for individual refinement, enabling more precise and effective optimization.

\begin{figure*}[t]
  \vspace{-2ex}
  \includegraphics[width=\linewidth]{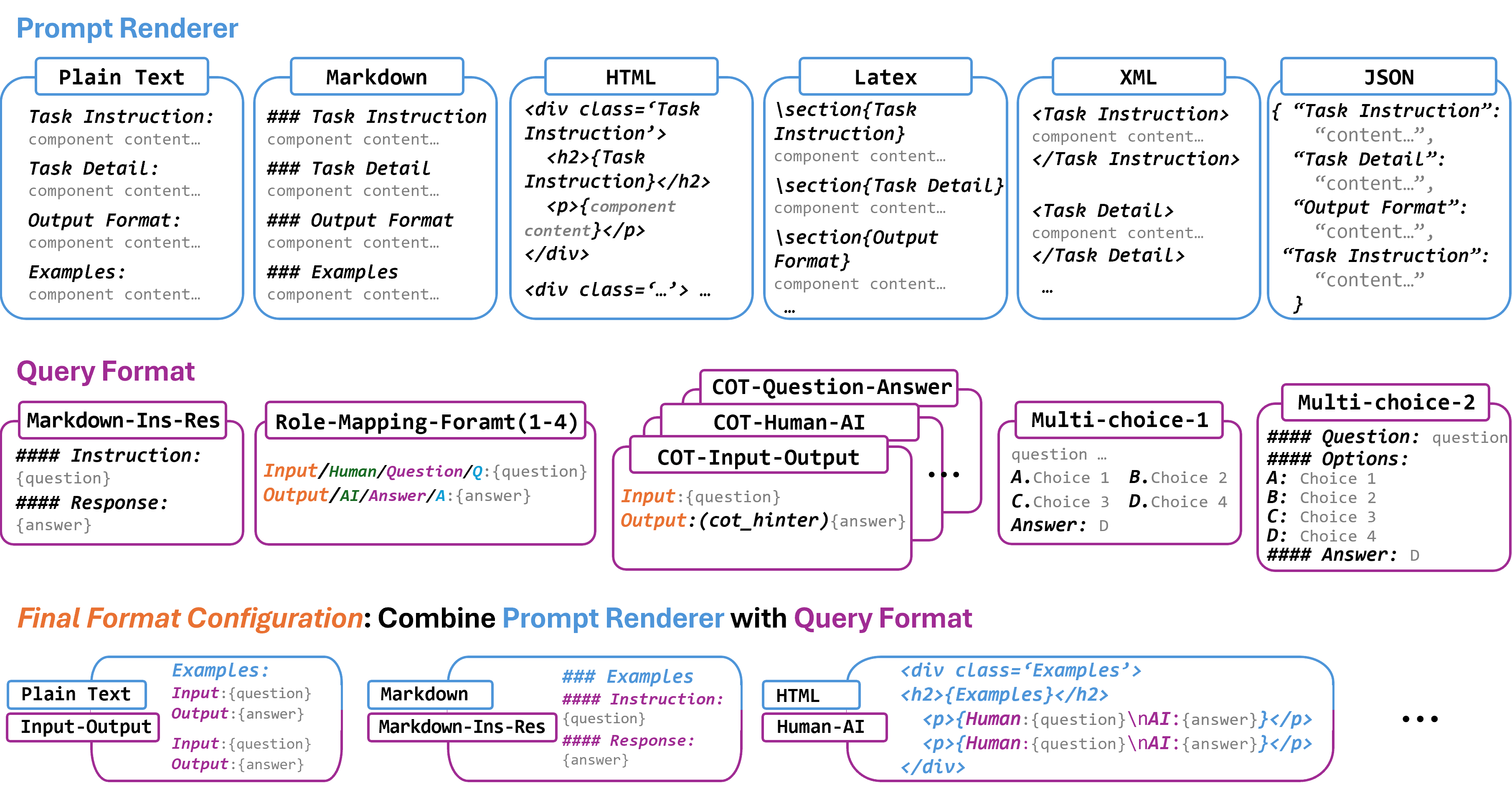}
  \vspace{-4.5ex}
  \caption {Built-in formats and rendering effects in our initial format pool. The final format configuration is achieved by selecting and combining elements from both the \textit{Prompt Renderer} and the \textit{Query Format} categories.}
  \vspace{-2ex}
  \label{fig:init_format}
\end{figure*}

\subsection{Format Optimization Design}
\label{sec:format_optimizer}

To efficiently navigate the prompt format space, \sysname{} incorporates a format optimizer based on a dynamic format pool and an LLM-assisted generation module. The optimizer iteratively explores, evaluates, and refines formatting choices.

\subsubsection{Format Pool with Scoring System}

The format pool maintains a diverse collection of prompt format configurations, organized hierarchically to represent variations at both macro (e.g., overall prompt structure) and micro (e.g., component-internal rendering) levels. In this work, we prototype this structure with two distinct renderers: a \textit{Prompt Renderer} defining the global structure, and a \textit{Query Format} governing the rendering of individual components, such as examples and queries, as shown in Figure~\ref{fig:init_format}.

Each format $f$ in the pool is associated with a performance score, $Q(f)$, which cumulatively reflects its effectiveness across different prompt contents.  When a format $f$ is evaluated on a new content $c$, its score is updated as follows: $Q(f) \gets Q(f) + m(c, f)$, where $m(c, f)$ is a task-specific metric function. For example, in reasoning tasks, $m(\cdot)$ represents the average accuracy (0 or 1) on an evaluation dataset.  The number of evaluations for each format, $N(f)$, is also tracked to enable score normalization and fair comparison.

To initiate the format exploration process, we define an initial format search space, $\mathcal{F}$, consisting of a set of predefined and commonly used formats (Figure~\ref{fig:init_format}), along with diverse variations that introduce subtle changes (e.g., spacing, punctuation, special symbols). This initial format pool serves as the foundation for subsequent exploration.

\subsubsection{LLM-assisted Format Generation}

To overcome the limitations of a static format pool and promote continuous adaptation, we introduce an LLM-based format generator, $LLM_{f\_gen}$. This module autonomously generates novel formats by leveraging information about the existing formats in the pool and their performance characteristics. By dynamically creating new formats, we aim to escape local optima and discover potentially superior formatting strategies.

In each optimization round, the format generator is provided with the existing formats and their normalized performance scores, $\frac{Q(f)}{N(f)}$, and tasked with creating new formats or variations that might improve performance. This iterative process not only diversifies the format pool but also ensures that our system can adapt to and incorporate a wide range of formats, thereby enhancing its utility and effectiveness.
Additional details and generated examples about the format generation process are available in Appendix \ref{apx:format_gen_meta} and Appendix \ref{apx:format_gen_and_examples}.

\subsubsection{Search Format via Format Optimizer}

\begin{algorithm}[t]
    \caption{Searching Optimal Format Given a Prompt Candidate}
    \label{alg:search_format}
    \renewcommand{\algorithmicrequire}{\textbf{Input:}}
    \renewcommand{\algorithmicensure}{\textbf{Output:}}
    \begin{algorithmic}[1]
        \REQUIRE $p_0=(c_0,f_0)$: initial prompt, $p=(c, \cdot)$: current prompt candidate(with content $c$), $\mathcal{F}$: dynamic format pool, $k$: number of formats, $m(\cdot)$: evaluation metric, $\mathcal{D}$: evaluation data.
        \STATE Initialize: $Q(f) \gets m(c_0,f)$, $N(f) \gets 1$ for all $f \in \mathcal{F}$

        \STATE \textbf{Format Selection}: $\mathcal{F}_{select} \gets \{f \in \mathcal{F}: f$ is in the top $k$ w.r.t. $UCT(f)\}$
        
        \STATE \textbf{Format Generation}:
        \FOR{each $i=0,1,...,k$}
            \STATE Generate format: $f_{new} \gets LLM_{f\_gen}(\mathcal{F})$
            \STATE Collect $f_{new}$ to $\mathcal{F}_{gen}$, and add $f_{new}$ to $\mathcal{F}$
        \ENDFOR
        
        \STATE \textbf{Format Evaluation}:
        \FOR{each $f \in \mathcal{F}_{select} \cup \mathcal{F}_{gen}$}
            \STATE Evaluate $m(c, f)$ with dataset $\mathcal{D}$
            \STATE $Q(f) \gets Q(f) + m(c, f)$
            \STATE $N(f) \gets N(f) + 1$
            \STATE Update $UCT(f)$ by Eq.~\ref{eq:uct}
        \ENDFOR
        \STATE $\hat f \gets \arg\max_{f \in \mathcal{F}_{select} \cup \mathcal{F}_{gen}}m(c,f)$
        \ENSURE The optimal format $\hat f$ for content $c$
    \end{algorithmic}
\end{algorithm}

For each content candidate generated by the content optimizer, the format optimizer aims to identify the most appropriate format from format pool. 
To navigate the balance between exploring new formats and exploiting known effective ones, we implemented the Upper Confidence Bounds applied to Trees (UCT) algorithm~\citep{kocsis2006bandit}. The UCT algorithm employs a selection criterion given by:
\begin{align}
UCT(f)=\frac{Q(f)}{N(f)} + \alpha \sqrt{\frac{\sum_f N(f)}{N(f)}}\label{eq:uct}
\end{align}
where $\alpha$ serves as a balancing hyper-parameter, adjusting the trade-off between exploration and exploitation.

The overall process, outlined in Algorithm~\ref{alg:search_format},
selects $2k$ formats for evaluation in each optimization round: $k$ existing formats from the pool (based on UCT score), and $k$ new formats generated by the $LLM_{f\_gen}$. The selected formats from both the existing pool ($\mathcal{F}_{select}$) and the newly generated pool ($\mathcal{F}_{gen}$) are then evaluated using a predefined metric function $m(\cdot)$, and the best-performing format among the tested candidates will be identified. The result is then incorporated into the pool for future iterations. By iteratively evaluating formats, the format optimizer ensures a balance between exploring new formats and refining current ones, converging to the best format configuration.



\subsection{Integrated Optimizer Design}
\label{sec:dual_optimizer}




\sysname{} integrates content and format optimization within an iterative loop, as illustrated in Figure~\ref{fig:pipeline}. In each iteration, the content optimizer proposes candidate prompts, and the format optimizer identifies the most effective format for each candidate. This ensures that each prompt benefits from the most effective formatting. Furthermore, effective formats can improve the diversity and performance of content candidates, thereby helping content optimizer for beam search in the next iteration.

In summary, the content and format optimizers work in tandem, leveraging the LLM to enable rapid adaptation and customization. This iterative and collaborative process optimizes both the content and the format of prompts, leading to significant improvements in prompt quality.

\begin{table*}
  \centering
  {\footnotesize
  \begin{tabular}{lcccc}
    \hline
    \toprule
    \textbf{Method} & \textbf{Mistral-7B-v0.1} & \textbf{LLaMA-3.1-8B} & \textbf{LLaMA-3-8B-Instruct} & \textbf{Phi-3-Mini-Instruct} \\
    \hline
    \rowcolor[gray]{0.95}\multicolumn{5}{c}{\rule{0pt}{2.5ex}\textbf{\textit{Big-Bench Classification}}\rule{0pt}{2.5ex}}\\
        Baseline  & 56.00          & 64.00          & 70.00          & 54.00      \\
        GRIPS           & 86.00          & 67.00          & 84.00          & 69.00      \\
        APE             & 73.00          & 65.00          & 60.00          & 63.00      \\
        ProTeGi             & 83.00          & 81.00          & 82.00          & 76.00      \\
        SAMMO           & 86.00          & 80.00          & 86.00          & 78.00      \\
        \textbf{\sysname{}} (\textit {Ours})            & \textbf{94.00} & \textbf{90.00} & \textbf{91.00} & \textbf{87.00} \\
    \hline
    
    \rowcolor[gray]{0.95}\multicolumn{5}{c}{\rule{0pt}{2.5ex}\textbf{\textit{ARC-Challenge}}\rule{0pt}{2.5ex}}\\
        Baseline    & 67.15          & 73.81            & 75.94          & 84.39 \\
        GRIPS               & 77.05          & 77.90            & 79.61          & 87.46 \\
        APE                 & 75.85          & 77.05            & 78.67          & 87.63 \\
        ProTeGi             & 76.54          & 77.22            & 79.86          & 87.54 \\
        SAMMO               & 77.22          & 77.13            & 79.86          & 87.03 \\
        \textbf{\sysname{}} (\textit {Ours})      & \textbf{79.35} & \textbf{78.50}   & \textbf{80.63} & \textbf{88.23} \\
    \hline

    \rowcolor[gray]{0.95}\multicolumn{5}{c}{\rule{0pt}{2.5ex}\textbf{\textit{GSM8K}}\rule{0pt}{2.5ex}}\\
        Baseline (1-shot cot)   & 36.85         & 50.03         & 74.00             & 83.45  \\
        Baseline (8-shot cot)   & 38.21         & 51.02         & 73.46          & 85.75 \\
        GRIPS                   & 39.04             & 50.27             & 74.53             & 83.47 \\
        APE                     & 40.33         & 52.39         & 75.13         & 83.85 \\
        ProTeGi                     & 45.72         & 54.74         & 75.36         & 84.84 \\
        SAMMO                   & 43.82         & 54.74         & 75.89         & 84.76 \\
        \textbf{\sysname{}} (\textit {Ours})                    &\textbf{53.22} &\textbf{63.38} & \textbf{80.74} & \textbf{89.16} \\
    \hline
    
    \rowcolor[gray]{0.95}\multicolumn{5}{c}{\rule{0pt}{2.5ex}\textbf{\textit{MATH-500}}\rule{0pt}{2.5ex}}\\
        Baseline (1-shot cot) &  4.60          & 10.58          & 12.20          & 12.60  \\
        Baseline (4-shot cot) & 10.20          & 23.40          & 14.00          & 40.40  \\
        GRIPS                 & 13.40          & 15.80          & 23.60          & 10.80 \\
        APE                   & 11.60          & 12.80          & 22.80          & 30.60 \\
        ProTeGi               & 10.80          & 17.00          & 18.40          & 28.80 \\
        SAMMO                 & 12.20          & 15.40              & 25.80          & 42.40 \\
        \textbf{\sysname{}} (\textit {Ours})       & \textbf{14.80} & \textbf{26.99} & \textbf{33.33} & \textbf{44.20} \\ \bottomrule
  \end{tabular}
  }
  \vspace{-1ex}
  \caption{\label{tbl:mainresults}
    Main results of \sysname{} and state-of-the-art methods on four datasets.
  }
  \vspace{-2ex}
\end{table*}
\section{Experiments}


\subsection{Experimental Setups}

\noindent \textbf{Models.}
We selected four open-source Large Language Models (LLMs) as our primary evaluation targets, including two foundational models, Mistral-7B-v0.1 \citep{jiang2023mistral7b} and LLaMA-3.1-8B~\citep{grattafiori2024llama3herdmodels}, as well as two instruction-tuned models, LLaMA-3-8B-Instruct \citep{llama3} and Phi-3-Mini-Instruct~\citep{abdin2024phi3technicalreporthighly}.
For content mutation and format generation during the optimization process, we employed GPT-4 (2024-05-01-preview) \citep{openai2024gpt4technicalreport}.

\noindent \textbf{Datasets.}
Our evaluation benchmark was designed to comprehensively assess model performance across a range of task types and difficulty levels, emphasizing diverse query formats. Specifically, we employed the following tasks:
\begin{itemize}[itemsep=-4pt, topsep=0pt, leftmargin=*, itemindent=-\labelsep]
    \item \textbf{Classification}: The \textit{Implicatures} task from the Big-Bench benchmark~\citep{srivastava2023beyond}.
    \item \textbf{Multiple-choice}: ARC-Challenge \citep{clark2018thinksolvedquestionanswering} task.
    \item \textbf{Reasoning}: GSM8K~\citep{gsm8k} and MATH500~\citep{math,lightman2023let} requiring complex reasoning abilities.
    
\end{itemize}

\noindent \textbf{Baselines.} 
We compared \sysname{} with several commonly used and popular baselines.
GrIPS~\citep{prasad2023grips} performs syntactic phrase-level edits in instruction, representing a non-LLM-based optimization approach. APE~\citep{zhou2023ape} and ProTeGi~\citep{C_2023EMNLP_APO} both employ LLM to optimize prompt content, but differ in mutation strategy. APE adopts an instruction induction approach, while ProTeGi leverages test cases feedback with LLM to guide the mutation process.
SAMMO~\citep{C_2024EMNLP_sammo} introduces a structured framework that incorporates a preliminary format mutation strategy, which relies on random selection from a predefined format pool.
All methods were evaluated using consistent experimental configurations to ensure a fair comparison.
We also report common baseline prompts for reasoning tasks, including 8-shot for GSM8K and 4-shot for MATH500.

\noindent \textbf{Implementation Details.}
To evaluate the generated prompts, we sample subsets from the training split for each benchmark (sizes: 50, 500, 500, and 300 examples respectively). Beam search (budget of 8) is used during prompt mutations.
Each experiment was capped at 20 iterations. Early stopping was implemented, halting the process if performance plateaus. The number of prompt components the LLM could modify decreased linearly from 4 to 1 over the iterations (see Appendix \ref{apx:opt_procedure}). We start with a single in-context example without any further instruction as the initial prompt for each model and task, except for GrIPS which requires an initial instruction. Detailed parameter settings and procedure are in the Appendix \ref{apx:implement_details}.

\subsection{Main Results}
Table~\ref{tbl:mainresults} summarizes the performance of \sysname{} compared with several state-of-the-art methods across four datasets.
The results highlight the superior performance of \sysname{}, significantly outperforming the baseline prompt and competing methods.

The effectiveness of \sysname{} is particularly evident when compared to methods like GRIPS, which relies solely on phrase-level mutations, yielding only marginal improvements. This highlights the importance of iterative feedback for effective prompt refinement, a feature shared by ProTeGi, SAMMO, and \sysname{}.
Furthermore, \sysname{} incorporates integrated format optimization that contributes significantly to the performance gains, particularly for previously challenging tasks.


The benefits of \sysname{} are particularly pronounced in reasoning tasks, such as GSM8K and MATH, known for their sensitivity to prompt structure. While improvements are observed on both datasets, the impact is more noticeable on GSM8K compared to the more complex MATH dataset, suggesting that task difficulty can moderate the attainable performance gains from prompt optimization.

Beyond performance metrics, we further evaluate the stability of \sysname{} and the computational cost, as detailed in Appendix \ref{apx:results_analysis}. Examples of optimal prompts are detailed in Appendix \ref{apx:opt_promtps}.

\subsection{Ablation Study}

\begin{table}[t]
  \centering
  {
  {\footnotesize
  \begin{tabular}{c|l|c|c}
    \toprule
    \textbf{Task} & \multicolumn{1}{c|}{\textbf{Method}} & \textbf{Llama3.1} & \textbf{Llama3-Ins} \\
    \midrule
      \multirow{5}{*}{BBC}      & ProTeGi          &  81.00    &   82.00   \\
                                & \sysname{}$_{f}$  &  83.00   &   86.00     \\
              & \sysname{}$_{c}$   &  85.00    &   {85.00}   \\
                       & \sysname{}$_{c+f}$ &  {88.00}     &   89.00       \\
                     & \textbf{\sysname{}}         &  \textbf{90.00}    &   \textbf{91.00}   \\
    \midrule

    \multirow{5}{*}{GSM8K}   & ProTeGi          &  54.74    &   75.36  \\
                               &\sysname{}$_{f}$   &  52.46   &   76.65     \\
                               & \sysname{}$_{c}$   &  {58.07}   &   {77.71}     \\
                     & \sysname{}$_{c+f}$ & 61.94      &   79.30       \\
                     & \textbf{\sysname{}}         &  \textbf{63.38}    &   \textbf{80.74}  \\
   
    \bottomrule
  \end{tabular}
  }
  }
  \vspace{-1ex}
  \caption{Performance comparison of the full \sysname{} pipeline against ablated variants with format-only (\sysname{}$_f$), content-only (\sysname{}$_c$), and sequential content-then-format (\sysname{}$_{c+f}$) optimization. \sysname{}$_c$ also compared to ProTeGi~\citep{C_2023EMNLP_APO}.}
  \label{tab:ablation_pipeline}
  \vspace{-2ex}
\end{table}

\noindent \textbf{Impact of the CFPO Pipeline Components.}
To understand the contribution of each component in \sysname{}, we compared the full \sysname{} approach against the following ablated variants: (1) \sysname{}$_f$ (format-only): optimizes the prompt format while holding content constant, (2) \sysname{}$_c$ (content-only): optimizes the prompt content while holding format constant, and (3) \sysname{}$_{c+f}$ (sequential): performs iterative content optimization followed by a separate, single-step format optimization. We also include ProTeGi~\citep{C_2023EMNLP_APO} as a baseline. While \sysname{}$_c$ shares the goal of content optimization with ProTeGi, \sysname{}$_c$ incorporates correct cases in diagnosis and component-wise mutation as key inovations. 
Table~\ref{tab:ablation_pipeline} presents the results of this analysis. \sysname{}$_c$ significantly outperforms ProTeGi, demonstrating the effectiveness of our content optimization strategy. Furthermore, the ablated variants all underperform compared to the complete \sysname{} pipeline. These results underscore the interdependence of content and format in prompt optimization, highlighting the importance of joint optimization for best performance.

\begin{table}[t]
  \centering
  \resizebox{\linewidth}{!}{
  {\footnotesize
  \begin{tabular}{c|l|c|c}
    \toprule
    
    \textbf{Task} & \multicolumn{1}{c|}{\textbf{Method}} & \textbf{Llama3.1} & \textbf{Llama3-Ins} \\
    \midrule
    \rowcolor[gray]{0.95} \multicolumn{4}{c}{\textit{Impact of Format Generation}} \\
    \midrule
    \multirow{2}{*}{{BBC}}   & w/o Format Gen        & 88.00       &  {87.00} \\
                                    & with Format Gen       & \textbf{90.00}       & \textbf{91.00}  \\
    
    \hline
    \multirow{2}{*}{{GSM8K}} & w/o Format Gen        & 62.70       & 78.85  \\
                                    & with Format Gen       & \textbf{63.38}       & \textbf{80.74}  \\
    \midrule

    \rowcolor[gray]{0.95} \multicolumn{4}{c}{\textit{Different Format Selection Strategies}} \\
    \midrule
                    & Random       &   85.00       &   87.00      \\
      BBC          & UCT($\alpha=0$)    &  86.00        &   88.00      \\
                     & UCT(ours)    &  \textbf{90.00}        &   \textbf{91.00}      \\
    \hline
                      & Random          &  62.40        &   78.82      \\
      GSM8K          & UCT($\alpha=0$)    &   {63.23}      &   79.08      \\
                     & UCT(ours)    &  \textbf{63.38}        &   \textbf{80.74}      \\
    
    \bottomrule
  \end{tabular}
  }
  }
  \vspace{-1ex}
  \caption{Ablation of format generation and comparison of format selection strategies.}
  \label{tab:ablation_format_gen}
\end{table}

\noindent \textbf{Analysis of Format Generation.}
We compared the full \sysname{} approach against variant that uses format from initial format pool without using LLM for generation. As shown in Table~\ref{tab:ablation_format_gen}, the version with format generation consistently outperforms the variant using only the initial pool. This highlights the benefit of our format exploration mechanism in expanding the prompt space. (See Appendix \ref{apx:abla_format_gen_models} for the ablation study with different format generation models.)

\noindent \textbf{Analysis of Format Selection Strategies.}
We further examined the effectiveness of our UCT-based format selection strategy. We compared it against two baselines: random selection from the format pool and a greedy selection approach (equivalent to setting $\alpha=0$ in Eq. (\ref{eq:uct}), disabling exploration). Table~\ref{tab:ablation_format_gen} shows that \sysname{} with UCT-based selection consistently achieves the best performance across all settings, demonstrating the efficacy of balancing exploration and exploitation in format searching.

\begin{table}[t]
  \centering
  \resizebox{\linewidth}{!}{
  {\footnotesize
  \begin{tabular}{l|c|c|c|c}
    \toprule
    \textbf{Optimizer} & \textbf{Mistral} & \textbf{Llama3.1} & \textbf{Llama3-Ins} & \textbf{Phi-3} \\
    \midrule
     Qwen2.5-14B         & 50.49   &  58.76  &  80.14 &  88.48   \\
     GPT-4 (CFPO)  & 53.22  &  63.38   &  80.74 &  89.16    \\
   
    \bottomrule
  \end{tabular}
  }
  }
  \vspace{-1ex}
  \caption{Performance of \sysname{} using different optimizer models.}
  \label{tab:ablation_optimizer}
  \vspace{-3ex}
\end{table}

\noindent \textbf{Exploring Different Optimizer Models.}
While GPT-4 was used as the primary optimizer model for \sysname{}, we also investigated the feasibility of using a more accessible model
, Qwen2.5-14B-Instruct. Table~\ref{tab:ablation_optimizer} presents the performance of \sysname{} with Qwen2.5-14B-Instruct as the optimizer. The results are encouraging, demonstrating that \sysname{} can be effectively adapted to smaller-sized open-sourced models, albeit with some performance trade-offs, increasing its potential for broader applicability.

\section{Conclusion}



This paper introduces \fullsysname{}, an innovative methodology that concurrently optimizes both prompt content and format. 
By leveraging a structured prompt template, \sysname{} decouples these elements, enabling integrated optimization and addressing a significant gap in current research that largely overlooks the critical influence of format. 
Our results demonstrate the substantial significant influence of format on LLM performance, underscoring the necessity of a joint optimization approach. These findings emphasize the importance of integrating content and format considerations in prompt engineering. CFPO represents a significant advancement, empowering developers to design effective and robust prompts and unlocking the full potential of LLMs across diverse applications.

\noindent\textbf{Limitations}
While the proposed method demonstrates promising results, there are several limitations worth noting.
First, the effectiveness of the approach is task- and model-dependent. 
While the method generates promising prompts for specific tasks and models, it may not generalize as effectively to others—particularly tasks that are less sensitive to prompt structure or models that already possess strong reasoning capabilities, thereby limiting its broader applicability.
Moreover, the iterative nature of the optimization process, with multiple mutation strategies, introduces computational complexity, which could hinder scalability in resource-constrained environments.
Finally, we acknowledge the inherent difficulty in definitively separating content and format, especially in real-world scenarios where structured data can be embedded within content. Future work will focus on addressing these limitations, including exploring the application of \sysname{} to complex agentic tasks and improving the stability and scalability of the optimization process.

\bibliography{string,ref}

\appendix 
\newpage

\section{Appendix: Detailed Optimization Process and Meta-Prompts}\label{apx:meta_prompts}

\subsection{Meta-Prompt Header Setup}

At the beginning of the prompt, we introduce the task and provide a detailed explanation of the prompt’s components, followed by the current version of the prompt. Below is the structure of the meta-prompt header, where placeholders are denoted in [ALL CAPS]:

\begin{lstlisting}[basicstyle=\ttfamily\footnotesize\color{gray},]
I'm trying to write a prompt to [TASK INTENTION].
The current prompt consists of several key components, including:
[DESCRIPTION OF COMPONENTS]

The complete prompt is as follows:
"""[CURRENT PROMPT]"""
\end{lstlisting}

\subsection{Content Optimization}\label{apx:content_opt_meta}

\subsubsection{Case-diagnosis and Revision}

As described in Section~\ref{sec:dual_optimizer}, content optimization is achieved through an iterative process of case-diagnosis and feedback guided mutation. To facilitate this process, we utilize three distinct meta-prompts, each tailored to a specific task within content optimization.

\noindent \textbf{Case Diagnosis Meta-Prompt.} This meta-prompt analyzes the current prompt's performance against a set of test cases. It identifies areas for improvement and suggests specific modifications for the next iteration.

\begin{lstlisting}[basicstyle=\ttfamily\footnotesize\color{gray},]
[META PROMPT HEADER]

Upon evaluating the current prompt, this prompt gets the following examples wrong:
[INCORRECT CASES]

Meanwhile, this prompt gets the following examples correct:
[CORRECT CASES]

Please review the provided examples of correct and incorrect answers, and identify [NUM OF DIAGNOSED COMPONENTS] specific area for improvement in the prompts. Each suggestion should focus on A SPECIFIC segment of the prompt that needs optimization. For each suggestion, provide a comprehensive explanation that encapsulates all the evaluation results. If you believe the EXAMPLES segment needs improvement, you may suggest one example that can be added, removed, or altered to enhance the EXAMPLES segment based on the examples given. If you think there is no need for improvement, do not return any prompt segment.
Please encapsulate each suggestion using the following format:

<START>
<Prompt segment: [Segment name]>
[Suggestion goes here]
<END>
\end{lstlisting}

\noindent \textbf{Feedback Application Meta-Prompt.} Based on the diagnosis, this meta-prompt generates targeted textual changes to enhance the prompt's performance. It directly modifies the identified components of the prompt based on the feedback.

\begin{lstlisting}[basicstyle=\ttfamily\footnotesize\color{gray},]
[META PROMPT HEADER]

The existing [COMPONENT NAME] segment contains:
[CURRENT CONTENT FOR THE COMPONENT]

Here are some suggestions for improving the [COMPONENT NAME] segments: 
[GENERATED DIAGNOSES]

Based on the above information, I wrote [NUMBER OF GENERATED CONTENT] distinct and improved versions of the [COMPONENT NAME] segment within the prompt.
Each revised segment is encapsulated between <START> and <END>. In case this segment is an empty string, generate a suitable one referring to the suggestion.
The [NUMBER OF GENERATED CONTENT] revised [COMPONENT NAME] segments are:
\end{lstlisting}

\noindent \textbf{Feedback Application Meta-Prompt (for Examples).} This meta-prompt specifically handles the optimization of few-shot examples. It revises examples by adding, deleting, or modifying one single instances, ensuring that the in-context learning process is effective.

\begin{lstlisting}[basicstyle=\ttfamily\footnotesize\color{gray},]
[META PROMPT HEADER]

The existing EXAMPLES segment contains:
[CURRENT IN-CONTEXT EXAMPELS IN PROMPT]

Here are some suggestions for enhancing the EXAMPLES segment: 
[GENERATED DIAGNOSES]

Based on the above information, I have crafted [NUMBER OF GENERATED EXAMPLES] improved version of the EXAMPLES segment within the prompt. Each revision represents ONLY ONE of the following specific actions:
1. Addition: Incorporating one new example into the existing set.
2. Deletion: Eliminating one single example from the current set.
3. Modification: Changing the content of an example while maintaining its contextual relevance.
Please present the results without indicating which action was taken. Each refined EXAMPLES segment is marked by <START> and <END>.

The [NUMBER OF GENERATED EXAMPLES] revised EXAMPLES are:
\end{lstlisting}

\subsubsection{Monte-Carlo Sampling}

\noindent \textbf{Monte-Carlo Sampling Meta-Prompt} explores a wider range of semantically equivalent yet syntactically varied instructions, enhancing the chances of discovering more effective prompts.

\begin{lstlisting}[basicstyle=\ttfamily\footnotesize\color{gray},]
[META PROMPT HEADER]

Please create a different version of [COMPONENT NAME] segment without changing its semantic meaning. In case this segment is an empty string, generate a suitable one. The existing [COMPONENT NAME] segment contains:
[CURRENT CONTENT FOR THE COMPONENT]

The varied [COMPONENT NAME] segment is as follows:
\end{lstlisting}

\noindent \textbf{Monte-Carlo Sampling Meta-Prompt (for Examples)} refines few-shot examples by strategically adding, deleting, or modifying single instances to ensure their effectiveness.

\begin{lstlisting}[basicstyle=\ttfamily\footnotesize\color{gray},]
[META PROMPT HEADER]

The existing EXAMPLE set contains:
[CURRENT IN-CONTEXT EXAMPELS IN PROMPT]

Please generate a variation of the EXAMPLES set within the prompt while keeping the semantic meaning. The revision shoud represent ONLY ONE of the following specific actions:
1. Addition: Incorporating one new example into the existing set.
2. Deletion: Eliminating one single example from the current set.
3. Modification: Changing the content of an example while maintaining its contextual relevance.
Please present the results without indicating which action was taken. The varied EXAMPLES segment is as follows:
\end{lstlisting}

\subsection{Format Generation}\label{apx:format_gen_meta}

Our format generation process is a two-step procedure designed to create diverse and effective prompt formats. We focus on generating two key components of a prompt's format: the \textit{Prompt Renderer} and the \textit{Query Format}. The appendix presents examples of the format generated using this pipeline.

\noindent \textbf{Step 1: Format Description Generation.} For each component (i.e., \textit{Prompt Renderer} and the \textit{Query Format}), we first generate a natural language description of the format, alongside an example of how this format would render a sample input. This description acts as a blueprint, guiding the subsequent code generation. We utilize a meta-prompt to instruct an LLM to perform this task. The meta-prompt takes existing format examples as context and generates new format descriptions along with rendered results. As an illustrative example, here is a conceptual outline of the meta-prompt employed for generating new \textit{Query Format} descriptions:

\begin{lstlisting}[basicstyle=\ttfamily\footnotesize\color{gray},]
[META PROMPT HEADER]

We have some preset QUERY_FORMAT candidates, here are our whole search pool:
[ALL EXISTING QUERY FORMATS DESCRIPTION]

Here are two examples from our QUERY_FORMAT candidates as for your reference:
<Format name: Question-Answer>
[RENDERED EXAMPLE 1]

<Format name: Instruction-Response>
[RENDERED EXAMPLE 2]

Please generate ONE new format for the QUERY_FORMAT segment, its description and render the provided example using this new format. The new format could either be a completely new format or a variation of an existing format. 

If you choose to generate a completely new format, please ensure that the new format is conventional, structured, and aligned with commonly used query formats. Avoid overly creative or unconventional formats that deviate significantly from standard practices. The new format should be distinct from the existing formats. 

The variation can focus on two parts, CASING and SEPARATOR:

CASING refers to both the capitalization of the text (e.g., f(x) = x.title(), f(x) = x.upper(), f(x) = x.lower()) and the specific wording or phrasing used (e.g., changing "question" to "instruction" or "input"). 

SEPARATOR: the punctuation or symbols used to separate the question and answer, there are some candidates as for your reference {{'', ' ', '\\n', '--', ';\\n', ' ||', '<sep>', ' \\n', ':', '.'}}.

Note that focus solely on the format itself without altering the content of the question and answer. The format should remain focused on the existing structure (e.g., Question/Answer or Instruction/Response) without modifying the content or introducing any new sections. Avoid the use of underlines or any unconventional formatting styles among words. The format name should only include alphanumeric characters and underscores. Special characters such as `|`, `!`, `#`, `@`, and spaces should be avoided.

Please encapsulate the new query format using the following format:

<START>
<Format name: [format name]>
<Description: [format description]>
[The example rendered by the newly generated format]
<END>

\end{lstlisting}

\vspace{2ex}

\noindent \textbf{Step 2: Format Code Generation.} Based on the natural language description and rendered example produced in Step 1, we subsequently generate the corresponding code implementation of the new format. This code will be used by the system to render prompts according to the defined format. We again leverage a meta-prompt to instruct the LLM, this time to generate the executable code. As an illustrative example, here is a conceptual outline of the meta-prompt employed for generating the code representation of a new \textit{Query Format}:


\begin{lstlisting}[basicstyle=\ttfamily\footnotesize\color{gray},]
[META PROMPT HEADER]

We have some preset QUERY_FORMAT candidates, here are our whole search pool:
[ALL EXISTING QUERY FORMATS DESCRIPTION]

Here are two code implementations from our QUERY_FORMAT candidates as for your reference:
<Format name: Question-Answer>
<Renderer code>
[Question-Answer RENDERER CODE]
<Extractor code>
[Question-Answer EXTRACTOR CODE]

<Format name: Instruction-Response>
<Renderer code>
[Instruction-Response RENDERER CODE]
<Extractor code>
[Instruction-Response EXTRACTOR CODE]

Here is the example rendered by the new format:
[RENDERED RESULTS]

Please generate the code for this provided example based on the new QUERY_FORMAT. Ensure that both the renderer and extractor functions are included. The generated code should be plain Python code without any Markdown syntax or language identifiers such as ```python or '''python. Please output the code directly without any additional formatting. If you need to use any additional and specific packages, please import them in the code. Note that the generated functions should include properly indented blocks, so they can execute without errors. Note that the renderer function name should be query_renderer_{format_name} and the extractor function name should be query_extractor_{format_name}.

Please encapsulate the code using the following format:

<START>
<Format name: {format_name}>
<Description: {format_description}>
<Renderer code>
[Renderer code]
<Extractor code>
[Extractor code]
<END>
\end{lstlisting}

\section{Appendix: Experimental Setup}\label{apx:implement_details}

\subsection{Hyperparameter in Optimization Procedure}\label{apx:opt_procedure}

During content optimization, case-diagnosis and Monte Carlo sampling each generate 4 prompts per round. A set of 40 test cases is used, with 5 correct and incorrect cases leveraged for case-diagnosis.
During optimization, the number of prompt components (e.g., Task Instruction, Task Detail, Output Format, Few-shot Examples) the LLM could modify simultaneously was dynamically adjusted across all the iterations. The number of components that could be modified decreases linearly from 4 to 1 over the iterations, promoting broad exploration of the prompt space initially and fine-grained refinement in later stages.

For format optimization, 4 UCT-selected formats and 4 newly generated formats are used to generate new prompts. The coefficient in the UCT selection process $\alpha$ is set to $1e-3$.

Each experiment was capped at a maximum of 20 optimization iterations.  An early stopping criterion was implemented, halting the process if the performance did not improve for a specified number of consecutive iterations.

\subsection{Model Generation Parameters}

Key generation parameters for both the LLM optimizer (GPT-4) and the target evaluation models are detailed below.

\begin{itemize}[itemsep=-4pt, topsep=0pt, leftmargin=*, itemindent=-\labelsep]
\item \textbf{LLM Optimizer (GPT-4):} We utilized the following settings to generate and refine prompts: \texttt{top\_p=1.0}, \texttt{max\_tokens=4096}, \texttt{seed=42}, and \texttt{temperature=1.0}.

\item \textbf{Evaluation Models:} The target evaluation model, evaluated on the generated prompts, was configured with: \texttt{top\_p=0.1}, \texttt{max\_tokens=256}, \texttt{seed=42}, \texttt{temperature=0.0}, \texttt{repetition\_penalty= 1.0}, and \texttt{stop='\textbackslash n'}.  The \texttt{stop} parameter ensured the model ceased generation upon encountering a newline character.  All generation processes were implemented using the vLLM library~\citep{kwon2023efficient}.

\end{itemize}

\subsection{Evaluation Metric}

Performance across all tasks was evaluated using the \textit{exact match} metric. For direct answer tasks (BigBench-Classification and ARC-Challenge), the model's direct prediction was assessed for an exact string match with the ground truth. For Chain-of-Thought reasoning tasks (GSM8K and MATH500), the final numerical answer was extracted from the generated explanation and compared to the correct solution for an exact match.
Finally, the best-performing prompt identified on the evaluation set for each optimization method is reported on the corresponding test set.

\subsection{Initial Prompt}\label{apx:init}

\subsubsection{Big-Bench Classification}

\textit{Prompt Renderer: Directly Joint}

\noindent \textit{Query Format: Input-Output}

\begin{lstlisting}[basicstyle=\ttfamily\footnotesize\color{gray},]
Examples:
Input: Speaker 1: 'You do this often?' Speaker 2: 'It's my first time.'
Output: no

{{Query placeholder}}
\end{lstlisting}

\subsubsection{ARC-Challenge}

\textit{Prompt Renderer: Directly Joint}

\noindent \textit{Query Format: MultiChoice\_QA}

\begin{lstlisting}[basicstyle=\ttfamily\footnotesize\color{gray},]
You are a commonsense helper. I will provide several examples and a presented question. Your goal is to pick the most reasonable answer among the given options for the current question. Please respond with the corresponding label (A/B/C/D) for the correct answer.

Here are some examples:

Question: Forests have been cut and burned so that the land can be used to raise crops. Which consequence does this activity have on the atmosphere of Earth?
Choices:
A: It reduces the amount of carbon dioxide production
B: It reduces the production of oxygen
C: It decreases the greenhouse effect
D: It decreases pollutants in the air
Answer: B

{{Query placeholder}}
\end{lstlisting}

\subsubsection{GSM8K}

\textit{Prompt Renderer: Directly Joint}

\noindent \textit{Query Format: QA}
\begin{lstlisting}[basicstyle=\ttfamily\footnotesize\color{gray},]
Q: There are 15 trees in the grove. Grove workers will plant trees in the grove today. After they are done, there will be 21 trees. How many trees did the grove workers plant today?

A: There are 15 trees originally. Then there were 21 trees after some more were planted. So there must have been 21 - 15 = 6. The answer is 6.

{{Query placeholder}}
\end{lstlisting}

\subsubsection{MATH500}
\textit{Prompt Renderer: Directly Joint}

\noindent \textit{Query Format: Question-Answer}
\begin{lstlisting}[basicstyle=\ttfamily\footnotesize\color{gray},]
A chat between a curious user and an AI assistant. The assistant gives step-by-step solutions to the user's questions. In the end of assistant's response, a final answer is given in the format of "The answer is: <ANSWER>.".

Here are some examples:
Question: Let \[f(x) = \left\{
\begin{array}{cl} ax+3, &\text{ if }x>2, \\
x-5 &\text{ if } -2 \le x \le 2, \\
2x-b &\text{ if } x <-2.
\end{array}
\right.\]Find $a+b$ if the piecewise function is continuous (which means that its graph can be drawn without lifting your pencil from the paper).
Answer: Let's think step by step. For the piecewise function to be continuous, the cases must "meet" at $2$ and $-2$. For example, $ax+3$ and $x-5$ must be equal when $x=2$. This implies $a(2)+3=2-5$, which we solve to get $2a=-6 \Rightarrow a=-3$. Similarly, $x-5$ and $2x-b$ must be equal when $x=-2$. Substituting, we get $-2-5=2(-2)-b$, which implies $b=3$. The answer is: $a+b=-3+3=\boxed{0}$.

{{Query placeholder}}
\end{lstlisting}

\section{Appendix: CFPO Performance, Stability, and Cost Analysis}\label{apx:results_analysis}

\subsection{CFPO Performance on MMLU}

To further evaluate the Content-Format Integrated Prompt Optimization (CFPO) approach, we assessed its performance on the MMLU benchmark.  Specifically, we focused on high school history categories  (including \textit{high\_school\_european\_history}, \textit{high\_school\_us\_history} and \textit{high\_school\_world} \textit{\_history}) to evaluate CFPO's effectiveness on knowledge-based tasks.

Table~\ref{tab:mmlu} summarizes the MMLU performance, showcasing improvements achieved with CFPO, which reinforce the generalizability of CFPO, extending its demonstrated benefits from reasoning tasks (as presented in the main paper) to knowledge-based tasks. This expanded validation will be integrated into the revised manuscript.

\begin{table}[t]
  \centering
  \resizebox{\linewidth}{!}{
  {\footnotesize
  \begin{tabular}{c|l|c|c}
    \toprule
    
    \textbf{Task} & \multicolumn{1}{c|}{\textbf{Method}} & \textbf{Llama3.1} & \textbf{Llama3-Ins} \\
    \midrule
    \multirow{2}{*}{{MMLU}}   & Baseline               & 78.84        & 82.03   \\
                                    & CFPO                   & \textbf{81.74}        & \textbf{83.77}   \\

    \bottomrule
  \end{tabular}
  }
  }
  \vspace{-1ex}
  \caption{Ablation of format generation and comparison of format selection strategies.}
  \label{tab:mmlu}
\end{table}

\subsection{Impact of In-context Examples and Prompt Length}\label{apx:shot}

\begin{figure}[t]
  \includegraphics[width=\linewidth]{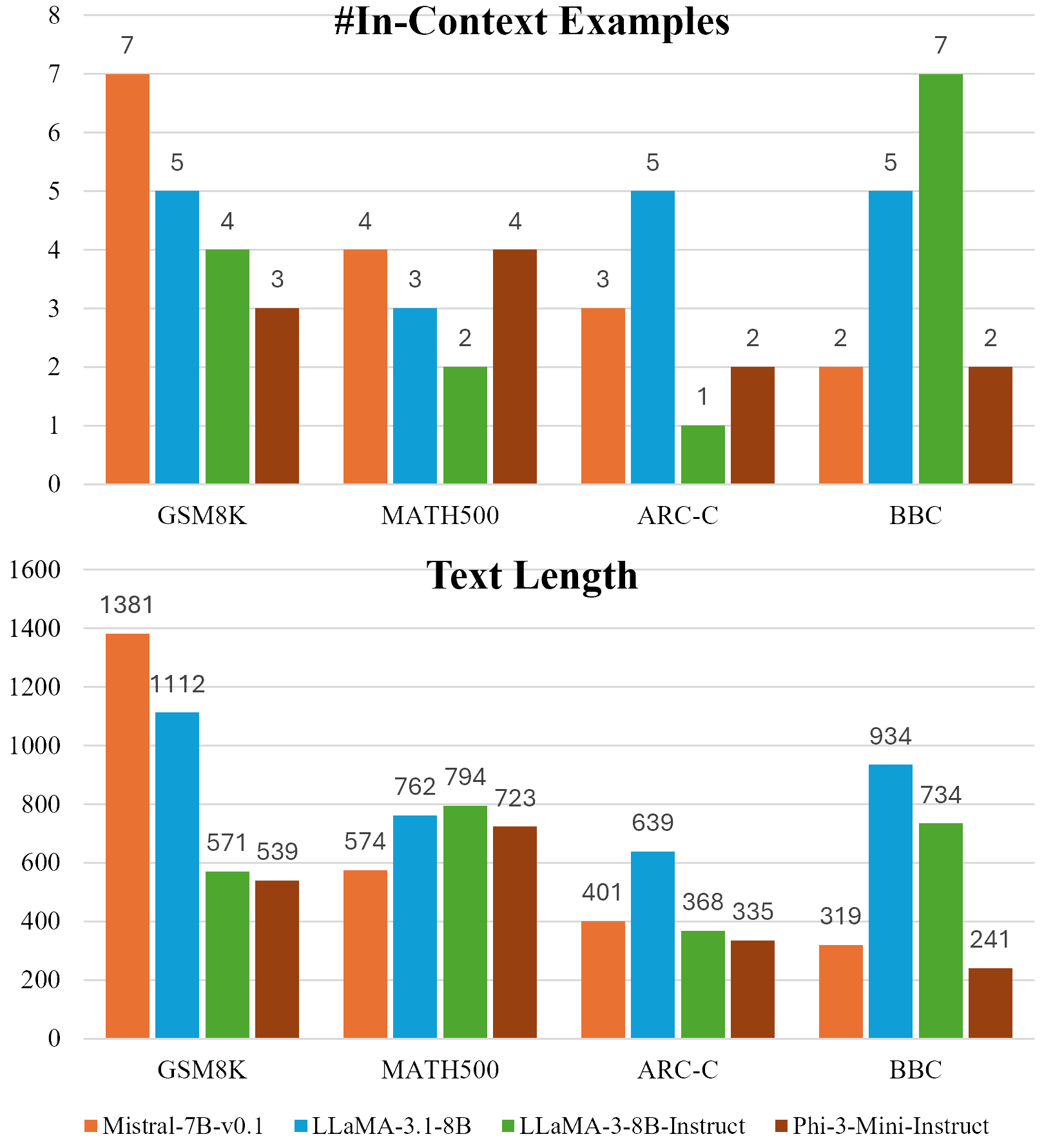}
  \caption {Overview of in-context examples and text lengths for optimized prompts on various tasks and models.}
  \label{fig:shot_results}
\end{figure}

Figure~\ref{fig:shot_results} presents an overview of the number of in-context examples and the text length of optimized prompts across various tasks and models. An interesting pattern emerges: pre-trained models consistently prefer prompts with longer text and more in-context examples compared to instruction-tuned models. 
This observation suggests that pre-trained models benefit more from explicit context and detailed reasoning steps, which align with their less task-specialized nature.
In contrast, the relative insensitivity of instruction-tuned models to prompt length and in-context examples supports the notion that these models have already trained with task-specific knowledge during fine-tuning, reducing their dependence on highly detailed prompts.

\subsection{Stability and Convergence Analysis}\label{apx:stability}

\begin{table}[t]
  \centering
  {
  {\footnotesize
   \begin{tabular}{lcc}
\toprule
\textbf{Model} & \textbf{CFPO} & \textbf{Reproduced CFPO} \\
\midrule
Mistral-7B-v0.1         & 53.22  & 51.87 ± 0.22 \\
Llama-3.1-8B            & 63.38  & 62.70 ± 0.59 \\
Llama-3-8B-Instruct     & 80.74  & 80.67 ± 0.17 \\
Phi-3-Mini-Instruct     & 89.16  & 88.89 ± 0.31 \\
\bottomrule
\end{tabular}}}
\vspace{-1ex}
    \caption{Stability Analysis: Comparison of reported CFPO scores with reproduced results (4 runs each).}
    \label{tab:stability_analysis}
\end{table}

\begin{figure}[t]
  \includegraphics[width=\linewidth]{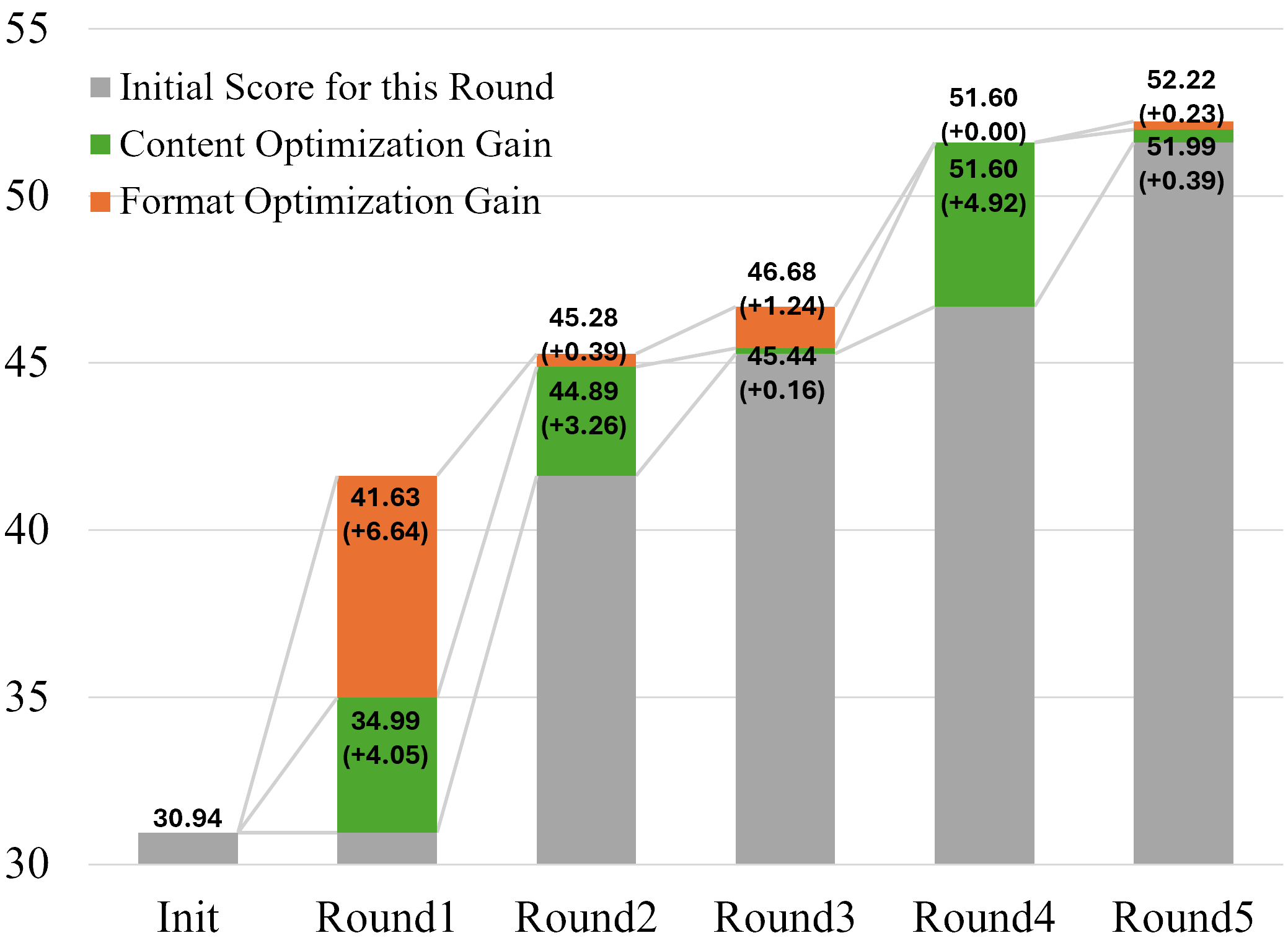}
  \vspace{-4ex}
  \caption {Convergence behavior of format and content optimization (Mistral-7B-v0.1 on GSM8K). Values indicate performance scores, with improvements from the previous sub-round in parentheses.}
  \vspace{-2ex}
  \label{fig:convergence}
\end{figure}

A crucial aspect of our framework is its stability. To quantify this, we performed multiple independent runs of CFPO and analyzed the variance in performance. Table~\ref{tab:stability_analysis} presents a comparison of the originally reported CFPO scores with the mean and standard deviation of scores obtained from four independent runs. The consistently low standard deviations across all models demonstrate the robustness and stability of our CFPO framework.

Furthermore, we examined the convergence behavior of content and format optimization.  Preliminary analysis using Mistral-7B-v0.1 on GSM8K (Figure~\ref{fig:convergence}) indicates that format optimization may converge faster than content optimization.  The performance gains in each sub-round (content or format optimization) are shown in parentheses. This suggests that the optimization schedule could be refined.

This difference in convergence speed could stem from the distinct optimization strategies: during content optimization, a diverse set of correct and incorrect examples is resampled from the training set in each round. In contrast, format optimization relies on generating variations of existing formats, leading to reduced diversity as the UCT score converges.  We hypothesize that allocating more resources to format optimization in early rounds, and subsequently shifting focus to content optimization, could improve overall efficiency. This is an area for further investigation in future work.

\subsection{Cost Analysis}\label{apx:cost}

\begin{table}[t]
  \centering
  \resizebox{\linewidth}{!}{
  {\footnotesize
    \begin{tabular}{lcccc}
    \toprule
    \textbf{Operation} &  \textbf{Tokens (I/O)} & \textbf{API Calls} & \textbf{Est. Cost}\\
    \midrule
    Case-diagnose & 2k/0.2k & 32 & \$ 0.83 \\
    Apply Feedback & 0.5k/0.1k & 110 & \$ 0.93 \\
    Gen Variation & 0.5k/0.2k & 81 & \$0.89 \\
    Gen Format & 0.5k/0.1k & 2 & \$0.016 \\
    Gen Format Code & 0.6k/0.2k & 2 & \$0.024 \\
    \midrule
    \textbf{Total (per Round)} & \textbf{159.6k}/\textbf{36.6k} & \textbf{237} & \textbf{\$2.69} \\
    \bottomrule
    \end{tabular}}}
    \vspace{-1ex}
    \caption{Average token usage, API calls, and estimated cost per round of CFPO optimization. \textbf{Tokens (I/O)}: typical input/output tokens per round, \textbf{API Calls}: average API calls per round, \textbf{Est. Cost}: estimated cost in \$.}
    \label{tab:cost_per_round}
    
\end{table}

\begin{table}[t]
  \centering
  \resizebox{\linewidth}{!}{
  {\footnotesize
    \begin{tabular}{lcc}
    \toprule
    \textbf{Task} & \textbf{Avg. Cost to Tar.} & \textbf{Avg. Time to Tar.} \\
    \midrule
    BBC & \$14.10 & 5.24 h \\
    ARC-C & \$16.81 & 6.25 h \\
    GSM8K & \$28.25 & 10.50 h \\
    MATH500 & \$19.85 & 7.38 h \\
    \bottomrule
    \end{tabular}}}
    \vspace{-1ex}
    \caption{Average cost and time required for CFPO to reach the reported target state-of-the-art performance on each benchmark.}
    \label{tab:cost_per_task}
    
\end{table}



An important consideration is the computational cost associated with \sysname{}. Table~\ref{tab:cost_per_round} breaks down the average token usage, API calls, and estimated cost per round of optimization for each operation within \sysname{}. These costs are calculated using current API pricing for the utilized LLMs.

Furthermore, \sysname{} incorporates an early stopping mechanism that terminates the optimization process when performance improvement plateaus.  This prevents unnecessary iterations and reduces overall cost. Table~\ref{tab:cost_per_task} details the average cost and time required to reach the reported performance on each benchmark task. As demonstrated in Table~\ref{tab:cost_per_task}, the average computational cost per task remains manageable.

\section{Appendix: Format Generation Examples and Ablation Study}

\subsection{Examples of Generated Format}\label{apx:format_gen_and_examples}

Here we select several format generated by GPT-4 in CFPO process.

\subsubsection{Query Format}

\noindent \textit{\textbf{Highlight\_Separator\_Case}}
\begin{lstlisting}[basicstyle=\ttfamily\footnotesize\color{gray},]
QUESTION > Statement 1 | Every element of a group generates a cyclic subgroup of the group. Statement 2 | The symmetric group S_10 has 10 elements.
OPTIONS > (A) True, True (B) False, False (C) True, False (D) False, True
ANSWER > C
\end{lstlisting}

\noindent \textit{\textbf{Cascading\_Statements}}
\begin{lstlisting}[basicstyle=\ttfamily\footnotesize\color{gray},]
Question: Statement 1 | Every element of a group generates a cyclic subgroup of the group. Statement 2 | The symmetric group S_10 has 10 elements.
Options:
  -A True, True
  -B False, False
  -C True, False
  -D False, True
   Answer: C
\end{lstlisting}

\noindent \textit{\textbf{QA\_Titlecase\_Separator}}
\begin{lstlisting}[basicstyle=\ttfamily\footnotesize\color{gray},]
Question || In 3 years, Jayden will be half of Ernesto's age. If Ernesto is 11 years old, how many years old is Jayden now?
Answer || Let's think step by step. Ernesto = 11 + 3 = <<11+3=14>>14 Jayden = 14/2 = <<14/2=7>>7 in 3 years Now = 7 - 3 = <<7-3=4>>4 Jayden is 4 years old.
\end{lstlisting}

\noindent \textit{\textbf{QA\_Brackets\_Colon\_Newline}}
\begin{lstlisting}[basicstyle=\ttfamily\footnotesize\color{gray},]
[Question]:
In 3 years, Jayden will be half of Ernesto's age. If Ernesto is 11 years old, how many years old is Jayden now?

[Answer]:
Let's think step by step.
Ernesto = 11 + 3 = <<11+3=14>>14 Jayden = 14/2 = <<14/2=7>>7 in 3 years Now = 7 - 3 = <<7-3=4>>4 Jayden is 4 years old.
\end{lstlisting}

\noindent \textit{\textbf{QA\_CapsBold\_ColonNewline}}
\begin{lstlisting}[basicstyle=\ttfamily\footnotesize\color{gray},]
**QUESTION**:
In 3 years, Jayden will be half of Ernesto's age. If Ernesto is 11 years old, how many years old is Jayden now?

**ANSWER**:
Let's think step by step.
Ernesto = 11 + 3 = <<11+3=14>>14 Jayden = 14/2 = <<14/2=7>>7 in 3 years Now = 7 - 3 = <<7-3=4>>4 Jayden is 4 years old.
\end{lstlisting}

\subsubsection{Prompt Renderer}

\noindent \textit{\textbf{Concise\_Bullet\_Points\_Renderer}}
\begin{lstlisting}[basicstyle=\ttfamily\footnotesize\color{gray},]
- Task Instruction: Write a function that returns the sum of two numbers.

- Task Detail: The function should take two numbers as input and return their sum.

- Examples: Input: 1, 2
Output: 3

- Query: Input: 1, 2
Output:
\end{lstlisting}

\noindent \textit{\textbf{Tabular\_Sections\_Renderer}}
\begin{lstlisting}[basicstyle=\ttfamily\footnotesize\color{gray},]
| Task Instruction | Write a function that returns the sum of two numbers. |
| Task Detail | The function should take two numbers as input and return their sum. |
| Examples | Input: 1, 2
Output: 3 |
| Query | Input: 1, 2
Output: |
\end{lstlisting}

\noindent \textit{\textbf{Checklist\_Format\_Renderer}}
\begin{lstlisting}[basicstyle=\ttfamily\footnotesize\color{gray},]
- [ ] **Task Instruction**
Write a function that returns the sum of two numbers.

- [ ] **Task Detail**
The function should take two numbers as input and return their sum.

- [ ] **Examples**
Input: 1, 2
Output: 3

- [ ] **Query**
Input: 1, 2
Output:
\end{lstlisting}

\subsection{Ablation Study of Format Generation Models}\label{apx:abla_format_gen_models}

\begin{table}[t]
  \centering

  {\footnotesize
    
\begin{tabular}{lc}
\toprule
\textbf{Format Generation Model} & \textbf{Accuracy} \\
\midrule
Claude-3.5-haiku          & 58.91 \\
Llama-3.3-70B-Instruct    & 61.71 \\
Gemini-2.0                & 62.09 \\
DeepSeek-R1                & 62.24 \\
GPT-4 (CFPO)          & \textbf{63.38} \\
\bottomrule
\end{tabular}}
    \caption{GSM8K Accuracy with Different LLMs for Format Generation (Using GPT-4 as content optimizer and LLaMA-3.1-8B as the target model).}
    \label{tab:format_gen_models}
\end{table}

To investigate the potential reliance on a specific LLM for format generation within our framework, we conducted an ablation study. This study evaluated the performance of several alternative LLMs in generating the formats used by CFPO, as shown in Table~\ref{tab:format_gen_models}. We systematically replaced the original format generation LLM (GPT-4) with alternative models: Gemini-2.0, DeepSeekR1, Claude-3.5-Haiku, and Llama-3.3-70B-Instruct. In these experiments, GPT-4 was consistently used for content optimization, and LLaMA-3.1-8B remained the target model for evaluating performance on the GSM8K benchmark.

The results demonstrate that the choice of LLM for format generation has a relatively minor impact on overall accuracy.  While variations in the specific generated formats are observed across different LLMs, the consistent performance suggests that CFPO's effectiveness is not critically dependent on a single LLM.  Specifically, as long as the LLM demonstrates the capability to generate variations and code based on the provided schema, the downstream performance on GSM8K remains stable.  This robustness suggests that the core principles of CFPO are transferable and not tied to the idiosyncrasies of a particular LLM. This suggests CFPO is not overly sensitive to the choice of format generation model.

\section{Appendix: Examples of CFPO Optimal Prompt} \label{apx:opt_promtps}

Here we selected several optimal prompts searched by \sysname{}.

\vspace{1ex}

\noindent \textit{\textbf{LLaMA-3.1-8B on ARC-C}}
\begin{lstlisting}[basicstyle=\ttfamily\footnotesize\color{gray},]
<div class='TaskInstruction'>
  <h2>TaskInstruction</h2>
  <p>Your mission is to meticulously assess each situation presented alongside a specific question, employing your critical thinking and analytical skills. Your task comprises not only identifying the most logical and coherent choice (A/B/C/D) but also thoroughly evaluating how each option connects or diverges from the question's essence. This requires a deep engagement with both the query and the choices, ensuring your reasoning is firmly anchored in the specifics of the options provided. It is essential to weave direct elements from the choices into your analysis, demonstrating a detailed understanding of how each option relates to the core question, and articulating why alternatives may be less fitting given the scenario. This approach ensures a nuanced and well-justified selection process, grounded in the interplay between the question context and the specific details of the available choices.</p>
</div>
<div class='TaskDetail'>
  <h2>TaskDetail</h2>
  <p>In addressing the questions set before you, it is imperative to delve deeper than mere superficial observations or initial judgments. Each scenario or question must be examined not just in its immediate context but within a broader spectrum, looking into the underpinning mechanisms or far-reaching effects of each option presented. This necessitates a thorough exploration of the larger implications and the scientific or logical foundations that dictate the outcomes. For instance, in environmental matters, it is vital to assess not just the immediate effects but the sustained impact on the ecosystem. In the realm of science, such as when discerning chemical processes, it is crucial to understand the molecular or atomic level changes that classify a reaction as a chemical change. This enhanced level of scrutiny and deeper analysis will lead to more accurate and well-founded choices, ensuring your responses are not just correct, but are also backed by a solid understanding of the underlying principles or long-term consequences.</p>
</div>
<div class='OutputFormat'>
  <h2>OutputFormat</h2>
  <p>For every query presented, your task is to identify the right choice from the options (A/B/C/D) accompanied by a concise rationale for your selection. This format is vital as it showcases the thought process leading to your decision, facilitating a comprehensive grasp and interaction with the task.</p>
</div>
<div class='Examples'>
  <h2>Examples</h2>
  <p>Here are some examples:

Question: Forests have been cut and burned so that the land can be used to raise crops. Which consequence does this activity have on the atmosphere of Earth?
A: It reduces the amount of carbon dioxide in the atmosphere
B: It reduces the availability of oxygen
C: It lessens the greenhouse effect
D: It lowers the levels of pollutants in the air
Answer: B

Question: What is the most critical practice to ensure electrical safety while operating devices?
A: Ensure the device does not come into contact with water.
B: Use the device with hands covered in oil.
C: Operate the device with wet hands.
D: Leave the device plugged in when not in use.
Answer: A

Question: Placing a plant cell in a hypertonic solution typically results in which of the following?
A: The cell expanding as it absorbs water.
B: No significant change due to the rigid cell wall.
C: The cell shrinking as water exits the cell.
D: Rapid division of the cell.
Answer: C

Question: What is the primary effect of using fossil fuels on global climate change?
A: It leads to a significant reduction in greenhouse gases.
B: It decreases the Earth's surface temperature.
C: It increases the amount of greenhouse gases in the atmosphere.
D: It contributes to a decrease in carbon dioxide levels.
Answer: C

Question: The process of photosynthesis in plants primarily involves which of the following transformations?
A: Converting oxygen and glucose into carbon dioxide and water
B: Transforming water and carbon dioxide into oxygen and glucose
C: Changing sunlight into chemical energy without producing oxygen
D: Producing carbon dioxide and glucose from oxygen and water
Answer: B

{{ query }}
\end{lstlisting}

\noindent \textit{\textbf{LLaMA-3.1-8B on GSM8K}}

\begin{lstlisting}[basicstyle=\ttfamily\footnotesize\color{gray},]
**Understanding the Task: A Foundation for Mathematical Problem-Solving**
Your task is to methodically analyze the information provided and logically deduce the correct answer to the mathematical problem. Delve into each relevant detail, ensuring no critical step or aspect is overlooked. Approach the solution with a detailed-oriented mindset, ensuring every part of the process is considered to arrive at an accurate conclusion. Reflect on all the elements that might influence your reasoning or calculation, striving for thoroughness in your analysis.

**Decoding Mathematical Language in Real-World Scenarios**
For the most effective problem-solving in mathematics, particularly when faced with intricate calculations over periods or under specific scenarios affecting results, an attentive and systematic method is key. Start by accurately determining the base numerical value. Then proceed by methodically listing every significant change whether it be increases, decreases, or modifications that impacts this base figure as the scenario unfolds, making sure to include each change in your overall computations. It's essential to focus on the concept of compounded operations, whether they're applied annually, monthly, or daily, and to thoughtfully evaluate the consequences of extraordinary events or circumstances (like an unexpected inheritance, a yearly loss, or a singular occurrence with a major impact) that might significantly shift the end calculations. Sharpen your attention on the dynamics of numerical relationships, particularly in cases involving ratios, proportions, and the impact of percentage changes over durations, to avoid common mistakes. Misunderstandings or misapplications of these numerical relationships can frequently cause inaccuracies. Thus, it is critical to scrutinize these mathematical relationships, whether they are of direct or inverse proportions, as well as the aggregate effects of consecutive percentage changes, as outlined in the problem description. This intensified attention is pivotal for an accurate and detailed resolution of complex issues, marked by multiplicative elements and interconnected circumstances. Reflect deeply on the significance of every step in the calculation process, absorbing the nuances of these changes, to systematically arrive at the most precise solution.

**Ensuring Your Solution Fits the Scenario Perfectly**
In presenting your solution, ensure it comprises both a numerical answer and a meticulously detailed explanation of the process leading to it. Begin with outlining the initial conditions and sequentially narrate the calculations you make at each step, highlighting any compounded operations or adjustments made to account for unique scenarios or conditions. This progression should clearly show how each step contributes to arriving at the final answer. For instance, if the task involves calculating the total costs saved over time with additional periodic benefits, your response should methodically explain: "Starting with an initial savings of X, plus Y every Z period, and considering an additional benefit of A every B period, leads to a total of...". This comprehensive breakdown not only bolsters the understanding of the mathematical principles applied but also provides a robust framework for identifying and rectifying any potential inaccuracies throughout the problem-solving process.

**Examples to Illuminate the Path**
To better grasp the concepts, consider the following illustrative examples:
Question: There are 15 trees in the grove. Grove workers will plant trees in the grove today. After they are done, there will be 21 trees. How many trees did the grove workers plant today? / ANSWER: Think through the problem step by step, diving into each segment for a thorough exploration to piece together the final answer. There are 15 trees originally. Then there were 21 trees after some more were planted. So there must have been 21 - 15 = 6. The answer is 6.

Question: A book club starts with a membership of 120. If the club increases its membership by 10% in the first year and then loses 5% of its members in the second year, what is the total membership at the end of the second year? / ANSWER: Think through the problem step by step, diving into each segment for a thorough exploration to piece together the final answer. The club starts with 120 members. In the first year, it increases by 10%, which is 0.10 * 120 = 12, so there are 120 + 12 = 132 members after the first year. In the second year, the club loses 5% of its members, which is 0.05 * 132 = 6.6, but since the number of members must be an integer, we consider a loss of 7 members (assuming the figure is rounded up for practical reasons). Therefore, there are 132 - 7 = 125 members at the end of the second year.

Question: Martin saves $10 every week. In addition, every third week, he earns an extra $15 from helping his neighbor. How much has Martin saved after 9 weeks? / ANSWER: Think through the problem step by step, diving into each segment for a thorough exploration to piece together the final answer. Martin saves $10 each week, so over 9 weeks, he saves 9 * $10 = $90. Additionally, every third week, he earns an extra $15, which occurs three times within 9 weeks (in the 3rd, 6th, and 9th weeks). So, he earns an extra 3 * $15 = $45 from helping his neighbor. Therefore, the total amount Martin has saved after 9 weeks is $90 + $45 = $135.

Question: A teacher divides a class into groups for a project. If the ratio of boys to girls in the class is 3 to 2, and there are 30 students in the class, how many boys are in the class? / ANSWER: Think through the problem step by step, diving into each segment for a thorough exploration to piece together the final answer. The total ratio units for boys to girls in the class is 3 + 2 = 5. With 30 students in the class, each ratio unit represents 30 / 5 = 6 students. Therefore, the number of boys, represented by 3 parts of the ratio, is 3 * 6 = 18. The answer is 18.

Question: Grandma wants to order 5 personalized backpacks for each of her grandchildren's first days of school. The backpacks are 20% off of $20.00, and having their names monogrammed on the backpack will cost $12.00 each. How much will the backpacks cost in total? / ANSWER: Think through the problem step by step, diving into each segment for a thorough exploration to piece together the final answer. The backpacks are 20% off of $20.00, so the price after the discount is $20.00 - ($20.00 * 20%) = $20.00 - $4.00 = $16.00 each. The monogramming costs an additional $12.00 per backpack. Therefore, the total cost for each backpack is $16.00 + $12.00 = $28.00. For 5 backpacks, the total cost will be 5 * $28.00 = $140.00. The correct answer is $140.00.

**Query**
{{query}}
\end{lstlisting}

\noindent \textit{\textbf{LLaMA-3-8B-Instruct on MATH-500}}
\begin{lstlisting}[basicstyle=\ttfamily\footnotesize\color{gray},]
- Task Instruction: A chat between a curious user and an AI assistant focused on solving mathematical and reasoning tasks. The assistant is expected to deliver step-by-step solutions to the user's questions, emphasizing mathematical accuracy and rigor throughout the process. It must ensure that each mathematical operation and logical deduction is carefully examined and validated to derive the correct solution. At the conclusion of the response, the final answer should be presented in the format of "The answer is: <ANSWER>.", thereby confirming the solution's validity and demonstrating a thorough understanding of the problem-solving approach.

- Task Detail: In addressing equation-based inquiries, precision in algebra, geometry, piecewise functions, complex numbers, and financial mathematics is paramount. This involves a detailed analysis of each equation, assessing every element and specific condition. For piecewise functions, it's critical to ensure continuity by solving for variables that maintain consistency across sections. In geometry, integrating measurements such as angles, lengths, and areas is fundamental. Algebraic queries require a consideration of all potential solutions and constraints, ensuring a comprehensive resolution. The addition of complex numbers into this mix necessitates a thorough understanding of their properties and operations to accurately determine both real and imaginary solutions. Similarly, tackling financial mathematics problems demands a deep comprehension of concepts such as compound interest, present value, and future value to make precise financial forecasts and comparisons. This holistic approach confirms that all aspects of the problem are considered and that the solution accounts for every requirement, assuring mathematical integrity in the resolution process.

- Output Format: 1. Solutions that involve fractions, square roots, or crucial mathematical figures (e.g., pi) must be simplified to their most fundamental form. This includes reducing fractions to their lowest terms and expressing square roots in their least complex radical form.
2. Avoid the use of decimals unless the question explicitly requires it or they are necessary for conveying the most precise value possible.
3. Present solutions involving square roots in their reduced radical form, ensuring the simplification process enhances comprehension without diluting mathematical integrity.
4. In scenarios involving complex numbers, represent answers in their standard form (a + bi), ensuring both 'a' and 'b' are presented in their simplest, most refined state. This emphasizes the need for a clear, coherent representation of solutions encompassing complex numbers.
5. Conclude your explanation with the statement: "The answer is: \[<ANSWER>\].", reinforcing consistency and clarity across various mathematical challenges. This concluding statement should encapsulate the solution in its simplest and most direct form, reflecting a thorough simplification and rationalization process.

Your explanation must delineate a detailed, step-by-step progression leading to the final solution. This approach is not merely about arriving at the correct answer but about illuminating the path taken to get there, ensuring a deep understanding and clear demonstration of the reasoning behind each step.

- Examples: Here are some examples:
### Instruction:
A rectangle ABCD has sides AB = 8 units and BC = 6 units. A circle with a radius r units is inscribed within this rectangle. Calculate the radius r of the inscribed circle, ensuring the answer is in its simplest form.

### Response:
We'll approach this problem by breaking it down into manageable steps. We start by understanding that the radius of the inscribed circle is equal to the distance from the center of the rectangle to any of its sides because the circle is perfectly inscribed. In a rectangle, this distance is half the length of the rectangle's shorter side. Therefore, the radius r of the inscribed circle is half the length of BC, which is $6 \div 2 = 3$ units. The answer is: $r=3$.

### Instruction:
Given a triangle where two sides are represented by complex numbers (3 + 4i) units and (1 - 2i) units, and the angle between them is 90 degrees, calculate the length of the hypotenuse. Ensure your answer includes a comprehensive breakdown of complex number operations and geometric principles applied.

### Response:
We'll approach this problem by breaking it down into manageable steps. We start by acknowledging that the length of a side represented by a complex number can be found using the modulus of that number. The modulus of the first side is $\sqrt{3^2 + 4^2} = 5$ units, and the modulus of the second side is $\sqrt{1^2 + (-2)^2} = \sqrt{5}$ units. Since these sides form a right triangle and we are given that the angle between them is 90 degrees, we can apply the Pythagorean theorem to find the length of the hypotenuse. The hypotenuse's length squared will be the sum of the squares of the lengths of the other two sides, which is $5^2 + (\sqrt{5})^2 = 25 + 5 = 30$. Thus, the length of the hypotenuse is $\sqrt{30}$ units. The answer is: $\sqrt{30}$.

- Query: 
{{query}}
\end{lstlisting}

\end{document}